\documentclass[preprint,12pt,nopreprintline]{elsarticle}

\usepackage{graphicx}
\usepackage{amssymb}
\usepackage{amsmath}
\usepackage{amsthm}
\usepackage{lineno}

\biboptions{comma,round,authoryear}

\begin{document}

\begin{frontmatter}
 
\title{Optimality Principles and Neural Ordinary Differential Equations-based Process Modeling for Distributed Control}

\author[First]{Michael R. Wartmann}
\author[Second]{B. Erik Ydstie}

\address[First]{Aletheia B.V., Arnhem, the Netherlands (e-mail: aletheia.io@pm.me).}
   
\address[Second]{Department of Chemical Engineering,
   Carnegie Mellon University, Pittsburgh, USA (e-mail: ydstie@cmu.edu).}

\begin{abstract}               
Most recent advances in machine learning and analytics for process control pose the question of how to naturally integrate new data-driven methods with classical process models and control. We propose a process modeling framework enabling integration of data-driven algorithms through consistent topological properties and conservation of extensive quantities. Interconnections among process network units are represented through connectivity matrices and network graphs. We derive the system's natural objective function equivalent to the non-equilibrium entropy production in a steady state system as a driving force for the process dynamics. We illustrate how distributed control and optimization can be implemented into process network structures and how control laws and algorithms alter the system's natural equilibrium towards engineered objectives. The basic requirement is that the flow conditions can be expressed in terms of conic sector (passivity) conditions. Our formalism allows integration of fundamental conservation properties from topology with learned dynamic relations from data through sparse deep neural networks.

We demonstrate in a practical example of a simple inventory control system how to integrate the basic topology of a process with a neural network ordinary differential equation model. The system specific constitutive equations are left undescribed and learned by the neural ordinary differential equation algorithm using the adjoint method in combination with an adaptive ODE solver from synthetic time-series data. The resulting neural network forms a state space model for use in e.g. a model predictive control algorithm.

\end{abstract}


\end{frontmatter}

\section{Introduction}
Modeling process systems poses the challenge of deriving a meaningful mathematical representation capturing the fundamental laws of nature while representing the systems actual real-world behavior within its given system context. A typical challenge within the context of process systems is that available data of the process is limited and individual systems differ significantly from each other. Hence, data of one system cannot be simply transferred or combined with data from similar other systems.  However, intelligent use of data in the process of modeling complex systems provides the context while model structure and general behavior can be derived from first principles knowledge of the system. 

A typical challenge within the context of process systems is that available data of the process is limited for the particular case at hand \cite{venkat}. Individual systems differ significantly from each other such that data of one system cannot be simply transferred or combined with data from similar other systems. Ontologies as in \cite{ontocape} can enable a link between data sets of similar systems in different contextual settings, however, structural alignment on their fundamental properties needs to take place first. As such, the need arises to model process systems in structurally consistent ways and take maximal advantage of the knowledge derived from first principles relationships.

While machine or deep learning algorithms are suitable to extract even very complex non-linear behavior in chemical engineering applications and mathematical programming \cite{leeml,fengqiML}, simple relationships such as fundamental mass, energy and component balances typically have to be learned ab initio, i.e., the data itself has to provide the context for those relationships. 

In \cite{schaefer}, model reduction approaches through machine learning are outlined and an artificial neural network (ANN) prototype for hybrid model structures is discussed. However, the authors state that reduced hybrid dynamic models will require many plant experiments to provide the necessary data for a high-fidelity model if no digital twin like plant monitoring is available. If however, initial fundamental laws can be provided as constraints, the information contained in the existing data could reduce the high-dimensional space by providing relationships naturally bounding the model to enhance fidelity of data-driven empirical relationships.  

To provide the most optimal starting point for any machine learning based model in the context of process systems, the structure of the material and energy flow of the process system would need to be embedded in the model itself. In deep learning, the backpropagation algorithm simply searches for the best possible fit of the given neural network structure to the input and output data provided. As a result, deep learning models have faced the challenge of explainability \cite{XAI}, i.e., parameters and structure of the resulting neural network cannot be explained or related to any fundamental laws of nature. ANN-based approaches would benefit therefore from understanding how connections between the sub units of a process network lead to complex system behavior and how these connections can be built into a such a model.

For the integration of data-driven methods and distributed control, we provide an organizational framework for process systems using ideas from network theory. The formalism of network theory has been particularly suitable for modeling and control of dynamic systems in electrical engineering applications. Developed in electrical circuit theory, the theory was extended to general thermodynamic systems by \cite{oster2,peusner}. 

The complexity of process systems arises from the variety of how simple sub units are connected \cite{hangos}. A crucial component in modeling process systems is therefore to understand how connections between the sub units lead to complex system behavior.  \cite{ya97} developed a theoretical framework providing a link between passivity theory and physics using the second law of thermodynamics. They discussed the need to develop passivity based control techniques which focus on input-output properties of the systems. An understanding for complex behavior can then be derived from macroscopic thermodynamic constraints instead of microscopic equations and the complexity that results from using very detailed models can be reduced. 
 
In this paper, we show how process networks can be regarded as gradient systems minimizing their own dissipation. Interconnections among process network units are represented through connectivity matrices and network graphs. A potential function is derived based on the so-called content and co-content of the process network. It is shown, how a process network minimizes its own entropy production, i.e., chooses the "path of least thermodynamic resistance" in the context of the second law as described in \cite{prigogine1947}. Properties can be altered through distributed control re-shaping the system's entropy dissipation based objective function as explained in Section \ref{sec:control}. This property introduces the notion of self-optimizing control \cite{skogestad} as the addition of a control loop can be understood in terms of altering the optimization objective. In Section \ref{sec:neural}, we derive a sparse neural network representation of the process system as a basis for e.g. ANN-based control. In a simple inventory control network example in Section \ref{sec:pipeline}, we demonstrate the principles.

\section{Process Networks}

Process networks can be written as a collection of interconnected sub-systems
\begin{eqnarray}
\label{eq:statespace}
\dot{x}_i&=&F(x_i) + \sum_{j=0, j \neq i}^{n}  G(u_i,x_i,x_j), \quad i=0,...,n\\
y_i & = & H(x_i)
\end{eqnarray}
$x_i$ is the state of subsystem $i$ and $x_i(0)$ is the initial condition. The function $F$ describes the unforced motion of the system, the function $G$ describes how the system is connected with other sub-systems, and the output function $H$ relates the state of the system to the measurement functions $y_{i}$. The functions $u_i$ represent the manipulated variables. The functions $F, G, H$ are all differentiable at least once. The state of the entire network is given by the vector $ x=(x_0^T,x_1^T,...,x_n^T)^T$.
 
Subscript zero refers to the reference (exo-) system. Often we are not interested in the dynamics of the exo-system, or more likely, it is too complex to model. The process system is modeled as the reduced system without the reference sub-system. Its state is given by the vector $ x=(x_1^T,...,x_n^T)^T$. The interactions with the exo-system are then established through the boundary conditions.

\begin{figure}[ht]
\centering
  \includegraphics[width=\textwidth]{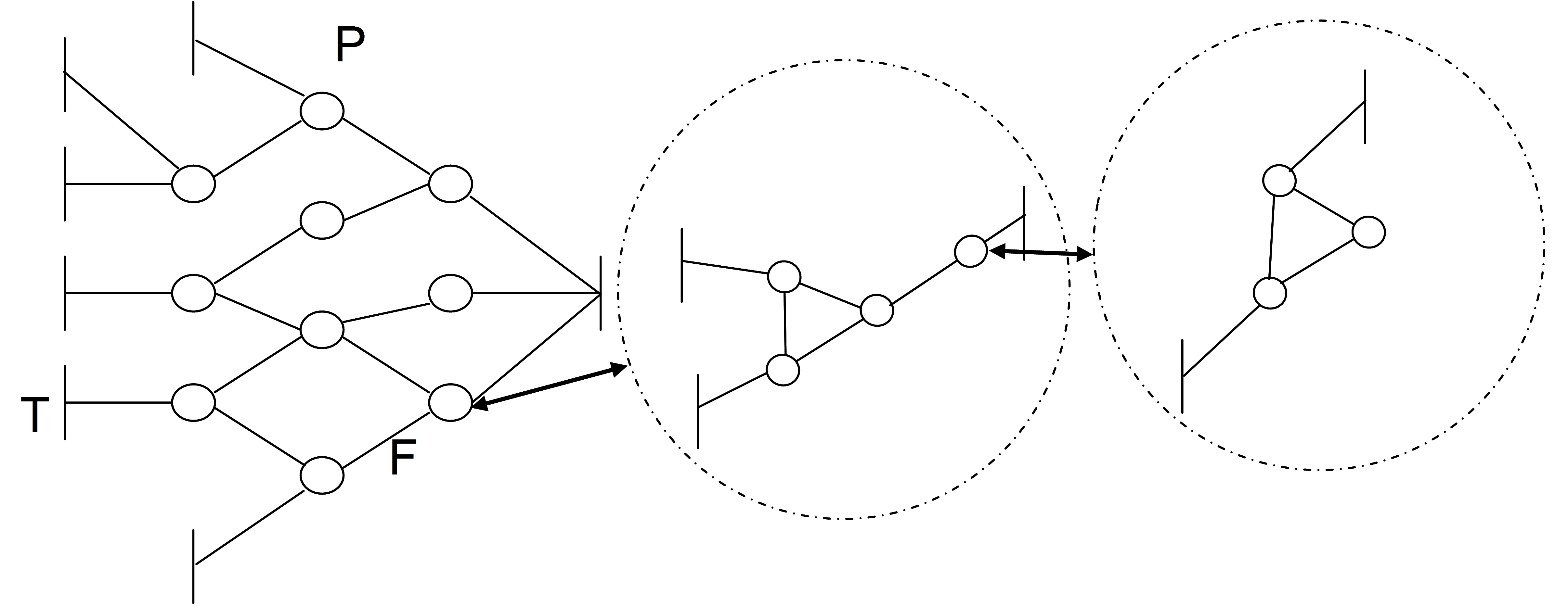}
  \caption{Graphical network representation: Topological structure of a network consisting of nodes, terminals, and flows. Nodes can contain sub-graphs and give rise to a hierarchical multi-scale structure.}
\label{fig:networkdef}
\end{figure}

The network form, as illustrated in Figure \ref{fig:networkdef} is convenient when we model systems with a graph structure. In such systems the interactions between the sub-systems depend on the state of the sub-system itself and the state of its immediate neighbors. Not all dynamical systems can be decomposed in this fashion. However, many large scale systems have sparse interconnections and they can be modeled compactly as networks of sub-systems with interconnections. It is also easy to see that many physical systems, especially those that satisfy the principle of local action, can be decomposed in the manner shown in (\ref{eq:statespace}).

We define the inventory $Z$ of a sub-system or a group of systems to be a non-negative, additive function of the state of the corresponding sub-system(s). By additivity we mean that if $Z_1$ is the inventory of sub-system 1 and $Z_2$ is the inventory of sub-system 2, then $Z_1+Z_2$ is the total inventory. Hence for any $i,j$
$$
Z\left(\begin{array}{c} x_i\\x_j\end{array} \right)= Z(x_i)+Z(x_j)
$$
By non-negativity we mean that the inventory cannot be less than zero. Examples of physical inventories include mass, energy and charge. More generally, an inventory is any property which is related to an amount. These properties allow us to conclude that the theory of process systems is related to the theory of positive systems described by \cite{luenberger79}.

By referring to (\ref{eq:statespace}) and using continuity we derive the conservation law
\begin{equation}
\label{eq:inventorybalance}
\frac{d Z_i}{dt} = p_i(x_i)+ \sum_{j=1, j \neq i}^{n} f_{ij}(u)
\end{equation}
The {\em drift} $p_i(x_i)=\frac{\partial Z(x_i)}{\partial x_i} F(x_i)$ measures the rate of production and the function $f_{ij}(u)=\frac{\partial Z(x_i)}{\partial x_i} G(u,x_i,x_j)$ measures the {\em supply} of $Z$ between sub-systems $j$ and $i$. We have the symmetry condition
$$
f_{ij}(u)=f_{ji}(u)
$$
The term
$$
\phi(u,z,d)=\sum_{j=1, j \neq i}^{n} f_{ij}(u)
$$
therefore measures the net rate of supply to sub-system $i$ from all other sub-systems. It is called the {\em action} on sub-system $i$.
\newdefinition{defn}{Definition}
\begin{defn}
Let $X_0$ be a subset of state-space. An action defined by (\ref{eq:inventorybalance}) is said to have the
\begin{enumerate}
\item  {\em Clausius-Planck property} if $p(x) > 0$ for $x$ not in $X_0$
\item  {\em Conservation property} if  $p(x)=0$ for all $x$ not in $X_0$
\item {\em Dissipation inequality} if  $p(x) < 0$ for $x$ not in $X_0$
\end{enumerate}
\noindent
The set $X_0$ associated with the dissipative action $\phi$ is called the set of {\em passive states}.
\end{defn}

\subsection{Network theory}

By a graph $\mathbf{G}$ we mean a finite set $\upsilon(\mathbf{G}) = (\upsilon_{1},...l,\upsilon_{n_{P}})$, whose elements are called \textbf{nodes}, together with the set $\epsilon(\mathbf{G}) \subset \upsilon \times \upsilon$, whose elements are called \textbf{branches}. A branch is therefore an ordered pair of distinct nodes.
\begin{itemize}
\item If, for all $(\upsilon_{i},\upsilon_{j}) \in \epsilon(\mathbf{G})$, the branch $(\upsilon_{j},\upsilon_{i}) \in \epsilon(\mathbf{G})$ then the graph is said to be \textbf{undirected}. Otherwise, it is called a \textbf{directed graph}.
\item A branch $(\upsilon_{i},\upsilon_{j})$ is said to be \textbf{incoming with respect to} $\upsilon_{j}$ and \textbf{outgoing with respect to} $\upsilon_{i}$ and can be represented as an arrow with node $\upsilon_{i}$ as its tail and node $\upsilon_{j}$ as its head.
\end{itemize}

\begin{defn}
A network of nodes $P_i, i=1,...,n_p,n_p + 1,...,n_{v}$ consisting of nodes and terminals interconnected through branches $E_i, i=1,...,n_{f}$ with topology defined by the graph
$$
\mathbf{G} = (\mathbf{E},\mathbf{P})
$$
\noindent
The system (\ref{eq:statespace}) is called a {\em process network} if its interconnection structure is described by a directed graph and we have
\begin{enumerate}
\item  {\bf First law:} There exists an inventory $E$ (the energy) which satisfies the conservation property
\item  {\bf Second law:} There exists an inventory $S$ (the entropy) which satisfies the Clausius-Planck property
\end{enumerate}
\end{defn}

We now develop a compact description of the topology of the network by introducing the incidence matrix. 
\begin{defn}
\label{def:incmatnode}
The $n_{t}\times n_{f}$ matrix $\mathbf{A_a}$ is called incidence matrix for the matrix elements $a_{ij}$ being
$$
a_{ij}=\left\{ \begin{array}{rl}
1, & \text{ if flow } j \text{ leaves node } i \\
-1, & \text{ if flow } j \text{ enters node } i \\
0, & \text{ if flow } j \text{ is not incident with node } i
\end{array}
\right.
$$ One node of the network is set as reference or datum node $P_0$ representing the exo-system. The $(n_{t}-1)\times n_{f}$ matrix $\mathbf{A}$, where the row that contains the elements $a_{0j}$ of the reference node $P_{0}$ is eliminated, is called reduced incidence matrix.
\end{defn}

The connections between nodes through branches can be uniquely defined using the incident matrix $\mathbf{A}$. The conservation laws (\ref{eq:inventorybalance}) can now be written
\begin{equation}
\mathbf{A} \mathbf{F}= \mathbf{0} \label{eq:KCLnode}
\end{equation} for the node-to-branch incident matrix $\mathbf{A}$, \\
where $ \mathbf{F^{T}} = [\frac{dZ_{1}}{dt},\frac{dZ_{2}}{dt},.., \frac{dZ_{n_{t}}}{dt},$ 
$f_{12},f_{13} .. f_{n_{t-1},n_{t}},p_{1},..,p_{n_{t}}]$. The flows $f_{ij}$ represent connections between two nodes i.e. $f_{ij}$ connects node $i$ to node $j$, $p_{i}$ denotes sources or sinks. The direction of the flows are defined according to the directionality established in the graph. We now define a vector $ \mathbf{W}$ so that
\begin{equation}
 \mathbf{W}= \mathbf{A^{T}} \mathbf{w} \label{eq:KVLnode}
\end{equation}
where $ \mathbf{W}$ are the potential differences across flow connections. The variables $w$ are conjugate to $Z$ if they are related via the Legendre transform of a convex potential like the entropy.

A dual structural representation can be derived using mesh analysis (the analysis developed above, which is based on the conservation laws, is called node analysis). Mesh analysis is counter-intuitive in process control applications but frequently used for electrical circuit analysis. When introducing stability and optimality concepts, we will focus on describing the primal problem only and its implications whereas the dual mesh-based problem holds equivalently (\cite{ADCHEM2009}).

\subsection{Constitutive Relations}

 Constitutive equations relate efforts and flows (resistive), flows and displacements (capacitive), and efforts and fluxes (inductive). The constitutive equations describe energy dissipating, irreversible processes (resistive) or energy storing, reversible processes. The constitutive equations define the type of energetic transaction inside the process system or between the process system and the environment. The three main types can be described as

\begin{itemize}
\item Capacitive constitutive equation: storage of potential energy, $W = f_{C}(Z)$
\item Inductive constitutive equation: storage of kinetic energy, $p = f_{L}(W)$
\item Resistive constitutive equation: dissipation of energy, $F = f_{R}(W)$
\end{itemize}

In the context of process networks, storage of energy usually occurs through capacitive elements. In this work, inductive constitutive equations are neglected due to the fact that we focus on chemical processes or chemical process plants in which inertial effects in mass flow and thus accumulation of kinetic energy are not a significant contributor to the energy balance.

\subsection{Graph-based Process Network Model}

The following set of equations defines the process system:

\begin{defn}
\label{def:networkeq}

\begin{eqnarray} \label{eq:system_network}
\mathbf{A} \mathbf{F} & = & \mathbf{0} \label{eq:KCLhere}\\
\mathbf{W}  & = & \mathbf{A^{T}}\mathbf{w} \label{eq:KVLhere} \\
\mathbf{F_{R}}  & = & \Lambda(\mathbf{W_{R}}) \label{eq:consteq} \\
\mathbf{Z}  & = & \mathbf{C}(\mathbf{w_{C}}) \label{eq:caplaw} \\
\mathbf{F_{R}} & = & \mathbf{F} - \mathbf{F_{S}} \label{eq:Fhat}\\
\mathbf{W_{R}} & = & \mathbf{W} - \mathbf{W_{S}} \label{eq:What}\\
\mathbf{F_{S}} & = & \mathbf{F_{T}} \label{eq:Fterminal}\\
\mathbf{W_{S}} & = & \mathbf{W_{T}} \label{eq:Wterminal0}\\
\mathbf{Z(0)} & = & \mathbf{Z_{0}} \label{eq:IC}
\end{eqnarray}
\end{defn}

The first two equations (\ref{eq:KCLhere}) and (\ref{eq:KVLhere}) are the Kirchhoff relations. Equations (\ref{eq:consteq}) are the resistive constitutive equations with $\Lambda$ being a matrix function and (\ref{eq:caplaw}) are the capacitive constitutive equations.

We introduced the variables $\mathbf{F_{R}}$ and $\mathbf{W_{R}}$ which facilitate writing the resistive constitutive equations in a compact way. Variables $\mathbf{F_{S}}$ and $\mathbf{W_{S}}$ are considered sources and hence interfaces to the external environment, the datum node, see Def. \ref{def:incmatnode}. The potentials $\mathbf{w_{C}}$ at the dynamic node, have been mathematically separated from the terminal potentials $\mathbf{w_{T}}$. The variables $\mathbf{F_{R}}$ and $\mathbf{W_{R}}$ are coordinate transformations which allow us to include the terminals as sources or sinks through (\ref{eq:Fterminal}) and (\ref{eq:Wterminal0}) for both, terminals where we have the function of the flows $\mathbf{F_{T}}$ or the potentials $\mathbf{W_{T}}$ given. For simplicity, we assume the terminal conditions as constant over time. The last equation (\ref{eq:IC}) constitutes the initial conditions for the inventories $\mathbf{Z}$.

The set of equations can be transformed to \ref{eq:dynsys3}, which is done solely for illustration purposes that \ref{eq:IC} is actually a system of nonlinear differential algebraic equations (DAE) 
\begin{eqnarray}
\frac{d \mathbf{Z}}{dt} &=& \mathbf{A(Z)} + \mathbf{B_{F}^{Z}(F_{T}^{input})} + \mathbf{B_{W}^{Z}(W_{T}^{input})} \label{eq:dynsys1first} \\
\mathbf{W_{T}^{output}} & = & \mathbf{C^{W}(Z)} + \mathbf{D_{F}^{W}(F_{T}^{input})} + \mathbf{D_{W}^{W}(W_{T}^{input})}  \label{eq:dynsys2} \\
\mathbf{F_{T}^{output}} & = & \mathbf{C^{F}(Z)} + \mathbf{D_{W}^{F}(W_{T}^{input})} + \mathbf{D_{F}^{F}(F_{T}^{input})}  \label{eq:dynsys3}
\end{eqnarray}

where non-linearities are only introduced through the constitutive equations. In this dynamic system, each terminal has an input and an output variable. The set of differential equations (\ref{eq:dynsys1first}) determines the trajectories of $\mathbf{Z}$ and represent a state space system. The algebraic constraints (\ref{eq:dynsys2}) and (\ref{eq:dynsys3}) compute the output variables at the terminals from the input variables and the state $\mathbf{Z}$.

To find the stationary solutions of the system, we need to solve the set of equations
\begin{eqnarray} \label{eq:system_network3}
\mathbf{A} \mathbf{F} & = & \mathbf{0} \label{eq:KCLhere2}\\
\mathbf{W}  & = & \mathbf{A^{T}}\mathbf{w} \label{eq:KVLhere2} \\
\mathbf{F} - \mathbf{F_{T}}  & = & \Lambda(\mathbf{W} - \mathbf{W_{T}}) \label{eq:consteq2}
\end{eqnarray} with the three main sets of constraints: Conservations laws, uniqueness conditions, and the constitutive equations. The inventories and capacitive constitutive equations are only relevant for the dynamic case.

\section{Self-optimizing Properties of Process Networks}

\cite{maxwell} formulated the minimum heat theorem which states that for linear resistive electrical circuits driven by constant power sources, the flows distribute themselves in a way as to minimize the heat that is dissipated through the resistive elements. \cite{prigogine1947} observed that the theorem can be generalized to general thermodynamic systems with the entropy production $\sigma_{S}$ being minimized at steady state. We can propose an optimization problem that allows us to find the steady state and dynamic trajectory of a process network.

\begin{defn}
For a process network with a graph $\mathbf{G}$, we can define the extended content
\begin{equation}
G =  \sum_{i=1}^{b} \int^{F_{i}} W_{i}dF_{i} = \int^{\mathbf{F}} \mathbf{W} d \mathbf{F} \label{eq:content}
\end{equation}
and the extended co-content:
\begin{equation}
G^{*} =  \sum_{i=1}^{b} \int^{W_{i}} F_{i}dW_{i} = \int^{\mathbf{W}} \mathbf{F} d \mathbf{W} \label{eq:cocontent}
\end{equation}
across all branches (also called edges) $b$. The extended content $G$ and co-content $G^{*}$ represent the sum of contents and co-contents for all branches i.e. reversible, irreversible, production and terminal flow connections of the network.
\end{defn}

In the following theorem, we introduce the connection between content, co-content and the Kirchhoff laws, and present how duality of the free variables  plays a crucial role for the optimization problem that is solved when a process network converges to a steady state solution. The constitutive equations are not directly involved as they are not relevant for the topological properties of the process network.
\newdefinition{thm}{Theorem}
\begin{thm}
\label{the:opt_node_KVL}
For the optimization problem
\begin{eqnarray}
\min_{\mathbf{w}} &  G^{*} = \int_{0}^{W} \mathbf{F} d \mathbf{W} \label{eq:pff1} \\
s.t. & \mathbf{W} = \mathbf{A^{T}} \mathbf{w} \label{eq:pff2}\\
     & \mathbf{F} =\Lambda(\mathbf{W}) \label{eq:pff3}
 \end{eqnarray} with the cocontent $G^{*}$ as objective function, the uniqueness conditions, and resistive constitutive equations as constraints, the solution exhibits a set of equations consisting of the uniqueness condition, the conservation laws, and the constitutive equations. The Langrange multipliers of the optimization problem are the network flow variables $\mathbf{F}$.
\end{thm}

\newdefinition{pf}{Proof}
\begin{pf}
Substituting the constitutive equations \eqref{eq:pff3} in \eqref{eq:pff1}, forming the Lagrangian and deriving the first order conditions show the result. (see \cite{ADCHEM2009} for a more detailed proof).
\end{pf}

In principle, an optimization problem is solved where one set of Kirchhoff equations is omitted. Through the first order conditions, the missing set of equations is derived. The optimization problem with the Kirchhoff voltage law as constraints can be converted to an optimization problem with the Kirchhoff current law and vice versa.

We can now propose the main theorem which allows us to connect the steady state of a process network to the objective function that is simultaneously optimized i.e. we can find the natural optimization problem that a process network solves, when converging to a steady state. We explored the structure of the problem in the previous theorem, however, we need to be able to define boundary conditions and a solution for  process networks connected to an exo-system.

\begin{thm}
\label{the:opt_probl}
Consider a process network $\mathbf{G}$ with given resistive constitutive equations $\mathbf{F_{R}} = \Lambda(\mathbf{W_{R}})$ and boundary conditions for each terminal as well as one set of either the conservation laws or the uniqueness conditions. The stationary solution ($\frac{dZ_{i}}{dt}=0$) for the network with conservation laws (\ref{eq:KCLhere}) and the uniqueness conditions (\ref{eq:KVLhere})

\begin{eqnarray} \label{eq:system_network2}
\mathbf{A} \mathbf{F} & = & \mathbf{0} \label{eq:KCLhere3}\\
\mathbf{W}  & = & \mathbf{A^{T}}\mathbf{w} \label{eq:KVLhere3} \\
\mathbf{F} - \mathbf{F_{T}}  & = & \Lambda(\mathbf{W} - \mathbf{W_{T}}) \label{eq:consteq3}
\end{eqnarray} can be found by solving the following optimization problem

\begin{eqnarray}
\min_{\mathbf{w}} &  G^{*} = \int_{0}^{W} \mathbf{F} d \mathbf{W} \label{eq:pf1}\\
s.t. & \mathbf{W} = \mathbf{A^{T}} \mathbf{w} \label{eq:pf2}\\
     & \mathbf{F_{R}} =\Lambda(\mathbf{W_{R}}) \label{eq:pf3}\\
     & \mathbf{F_{T}} = \text{const}  \text{ and/or } \mathbf{W_{T}} = \text{const} \label{eq:pf4}
 \end{eqnarray} or its equivalent dual optimization problem where (\ref{eq:KVLhere3}) is replaced by (\ref{eq:KCLhere3}).
\end{thm}

\begin{pf}
Substitution of the constitutive equations (\ref{eq:pf3}) into the objective function (\ref{eq:pf1}) to eliminate the flow variables $\mathbf{F}$. First order conditions of the resulting Lagrange function leads to equations (\ref{eq:KCLhere3}) - (\ref{eq:consteq3}).  See full proof in \cite{ADCHEM2009}.
\end{pf}

Concerning the second order conditions, we observe that convexity of the constraints is trivial for the linear Kirchhoff laws. Non-convexities of the optimization problem are due to non-linearities of the constitutive equations i.e. the constitutive equations are non-positive. For the second order conditions, it is apparent that the first derivative of the constitutive equations has to be analyzed and found positive definite for a global minimum, which corresponds exactly to the findings for passivity in \cite{krjpaper} for a unique network solution and convergence.

Generally, the objective function is a measure for dissipation of the storage variable over time. We conclude that the steady state of a passive network minimizes the dissipated power subject to the constraints imposed by the constitutive equations, topology, and boundary conditions, i.e. terminal connections.

\begin{thm}
The potential function of a process network is given as 
\begin{displaymath} 
P(\mathbf{Z}) = \Phi(\mathbf{Z}) + \int_{0}^{\mathbf{w_{T}}} \mathbf{F_{T}}^{T} d \mathbf{w_{T}}
\end{displaymath}
where $\Phi(\mathbf{Z}) = \int_{0}^{\mathbf{Z}} \mathbf{A_{R}} \Lambda[\mathbf{A_{R}}^{T} C^{-1}(\mathbf{Z})]^{T} d \mathbf{Z}$
\end{thm}

\begin{pf}
Starting with the cocontent function \eqref{eq:cocontent} and proceeding with the decomposition into terminals and resistive flows
\begin{equation}
G^{*} = \int_{0}^{\mathbf{W}} \mathbf{F}^{T} d \mathbf{W} = \int_{0}^{\mathbf{W_{R}}} \mathbf{F_{R}}^{T} d \mathbf{W_{R}} + \int_{0}^{\mathbf{w_{T}}} \mathbf{F_{T}}^{T} d \mathbf{w_{T}}
\end{equation} Using Kirchhoff's voltage law \eqref{eq:KVLhere} and the resistive \eqref{eq:consteq} and capacitive constitutive equations \eqref{eq:caplaw}, the result follows immediately.
\end{pf}

\section{Distributed Control and Optimization}
\label{sec:control}

In this section, we demonstrate, how the dynamics of the process network change under the assumption of perfect control \cite{garcia}. We assume that a process network can be optimized in an ideal manner, if all inventories can be controlled perfectly. 
\begin{defn}
An inventory control strategy is said to be {\bf direct}, if the assignment of the controllable flow $F_{K,i}$ and the inventory $Z_{i}$ are incident, i.e. the flow $F_{k,i}$ either enters (supply) or leaves the node (demand).
\end{defn} 
The concept of direct control is important, since it is related to the idea of passive systems.

\begin{figure} [ht]
\centering
  \includegraphics[width=\textwidth]{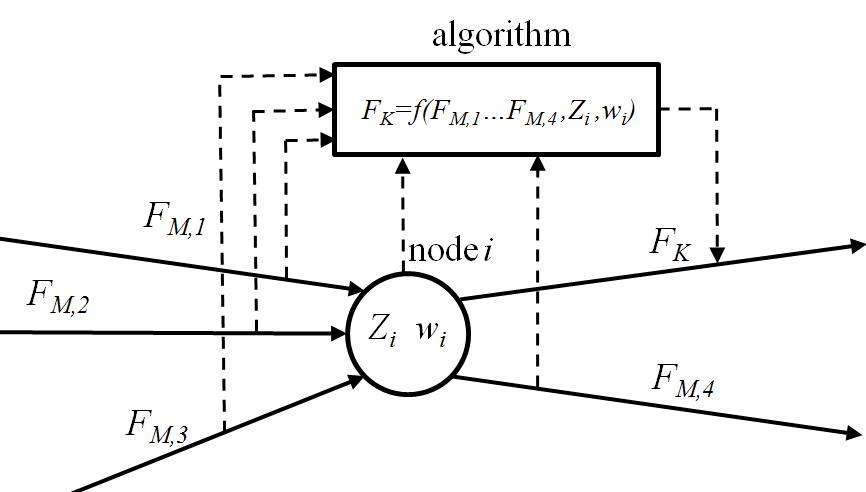}
  \caption{An algorithm measuring the flows $F_{M,i}$ and the potential $w_{i}$ or inventory $Z_{i}$ as inputs and computes the control action taken by the controlled flow $F_{K}$.}
\label{fig:algorithm}
\end{figure}

\begin{defn}
A process network with $n_{p}$ inventories $Z_{i}$ is controllable, if there exists one independent directly controlled flow $F_{k,i}$ for each $Z_{i}$.
\end{defn}
We define an algorithm for the direct control of an inventory $Z_{i}$ at node $i$ with the controllable flow $F_{K,i}$
\begin{equation}
\label{eq:controllaw}
F_{K,i} = \sum F_{M,j} - f(Z_{i}) 
\end{equation} where $F_{M,j}$ are the remaining flows incident with the node $i$. These flows have to be measured or related to measured quantities through models. $f(Z_{i})$ is a feedback function being linear, nonlinear, or discrete in order to describe how deviations from the optimal state are handled by the flow $F_{K,i}$. A graphical illustration of the information flow for the algorithm can be seen in Fig. \ref{fig:algorithm}.
 Using the decentralized proportional controller as an example, we get the gains of the proportional controllers as diagonal matrix $\mathbf{K_{C}}$ and $\mathbf{Z}^{c}$ is the vector of set-points for the inventory. An explicit expression for the control law follows from Kirchhoff's current law as

\begin{equation}
\label{eq:controllawlin}
\mathbf{F_{K}} = -\mathbf{A_{K}}^{-1} \mathbf{A_{R}} \mathbf{F_{R}} + \mathbf{A_{K}}^{-1} \mathbf{K_{C}}(\mathbf{Z}-\mathbf{Z^{c}})
\end{equation} where the matrix $\mathbf{A_{K}}$ describes the topology of the connections between the flows having an actuator $\mathbf{F_{K}}$ and the inventories $\mathbf{Z}$ targeted to be controlled by the actuator. The matrix $\mathbf{A_{R}}$ captures the connections between the nodes and the resistive flows with no control actuator $\mathbf{F_{R}}$. The matrix $\mathbf{A_{K}}$ is symmetric and full rank, hence invertible. 

\begin{thm}
\label{th:control}
The potential function of a process network $P(\mathbf{Z})$ with fixed potentials at the terminals leads to the following potential function through the application of proportional controllers \eqref{eq:controllawlin} for the controllable flows as
\begin{equation}
P^{c}(\mathbf{Z}) = \mathbf{K_{C} C}^{-1} \int^{\mathbf{Z}} (\mathbf{Z - Z^{c}})^{T} d \mathbf{Z} \label{eq:control}
\end{equation} 
\end{thm}

\begin{pf}
We begin with the optimization problem  \eqref{eq:pff1}-\eqref{eq:pff3}.
We split the objective function and constraints into subsets specific to the controllable flows $\mathbf{F_{K}}$, resistive flows $\mathbf{F_{R}}$, and terminal flows $\mathbf{F_{T}}$ 

\begin{eqnarray}
\min_{\mathbf{w}}   G^{*} &= \int_{0}^{\mathbf{W_{K}}} \mathbf{F_{K}}^{T} d \mathbf{W_{K}} + \int_{0}^{\mathbf{W_{R}}} \mathbf{F_{R}}^{T} d \mathbf{W_{R}} \nonumber  \\
&+ \int_{0}^{\mathbf{W_{T}}} \mathbf{F_{T}}^{T} d \mathbf{w_{T}} \label{eq:terms}
\end{eqnarray}
\begin{eqnarray}
s.t. &  \mathbf{W_{K}} = \mathbf{A_{K}}^{T} \mathbf{w} \\
    &   \mathbf{W_{R}} = \mathbf{A_{R}}^{T} \mathbf{w} \\
    &   \mathbf{W_{T}} = \mathbf{A_{T}}^{T} \mathbf{w_{T}} \\
   &   \mathbf{F_{K}} =\Lambda_{K}(\mathbf{W_{K}}) \\
    &  \mathbf{F_{R}} =\Lambda_{R}(\mathbf{W_{R}}) \\
   &   \mathbf{w_{T}}  =  \text{const.} \label{eq:Wterminal}\\
    &  \mathbf{Z}   =  \mathbf{C}\mathbf{w} \label{eq:capa}
\end{eqnarray} The constitutive equations for $\mathbf{F_{R}}$ are redundant, since we assume that their resistances are adjusted freely using the control law, given by \eqref{eq:controllawlin} and added to the optimization problem as constraints. The derivatives for the potential differences follow as $d \mathbf{W_{K}} = \mathbf{A_{K}} d \mathbf{w} $, $d \mathbf{W_{R}} = \mathbf{A_{R}} d \mathbf{w}$ and $d \mathbf{w_{T}} = \mathbf{0}$
 where $\mathbf{A_{K}}^{T} = \mathbf{A_{K}}$ since $\mathbf{A_{K}}$ is diagonal. The control law for $\mathbf{F_{K}}$ is now substituted into the objective function \eqref{eq:terms}. Using $\mathbf{A_{K}}^{-1} \mathbf{A_{K}} = \mathbf{I}$ and substituting \eqref{eq:Wterminal} and \eqref{eq:capa} into \eqref{eq:terms} we get
\begin{eqnarray}
\label{eq:final_obj}
\min_{\mathbf{Z}} &  P^{c}(\mathbf{Z}) =  \int_{\mathbf{0}}^{\mathbf{Z}} (\mathbf{Z}-\mathbf{Z^{c}})^{T} \mathbf{K_{C}} \mathbf{C}^{-1}  d \mathbf{Z}   \\
s.t.  & \mathbf{F_{R}} =\Lambda_{R}(\mathbf{W_{R}}) 
 \end{eqnarray} and the result follows. 
\end{pf} Effectively, this shows that the influence of the resistive constitutive equations is removed from the potential function and dominated by the control action in the ideal case. 

Under the assumption of constraints on the controller action in form of lower and upper bounds for the controlled flows $\mathbf{F_{K}^{LB}} \leq \mathbf{F_{K}} \leq \mathbf{F_{K}^{UB}}$, the system returns to its original dynamics if the controlled flows hit a boundary limit. Generally, the dynamic system with the original potential function
\begin{displaymath}
P(\mathbf{Z}) = \frac{1}{2} (\mathbf{Z} - \mathbf{Z}^{*})^{T} \Phi (\mathbf{Z} - \mathbf{Z}^{*})
\end{displaymath} changes into the control potential function
\begin{displaymath}
P^{c}(\mathbf{Z}) = \frac{1}{2} (\mathbf{Z} - \mathbf{Z}^{c})^{T} \mathbf{K_{C}} \mathbf{C}^{-1} (\mathbf{Z} - \mathbf{Z}^{c})
\end{displaymath} for the linear quadratic case. Theorem \ref{th:control} shows that it is possible to change the original steady state and dynamics of the process network to a desired control potential function under certain conditions.

In designing decentralized control schemes of large interconnected physical networks, some of the flows cannot be measured or controlled directly. Controlling or measuring all flows is in many cases not necessary or desired if the engineering objective aligns with the natural self-optimizing property (e.g. throughput maximization).

\section{Relationship of Process Network Description and Neural Networks}
\label{sec:neural}
A process network description based on graph theory as given in \eqref{eq:system_network} - \eqref{eq:IC} can be seen as a starting point for building a neural network model of a dynamic system with explainable parametrization (\cite{XAI}). In many applications, the point of departure is a fully connected neural network, i.e., fundamental relationships of the "to be learned" physical system have to be extracted from data and would require weights to be set to zero through training. In many applications, it would be desirable to explain parameters and structure in a neural net which is e.g. used for optimization and control such that models can be updated intuitively and understood from an operational point of view.

From an algorithmic point, we can regard the process network as a gradient system in its inventories or potentials. It is possible to design a neural network-based algorithm to compute the trajectory of a process system over time. For simplicity, we choose a set of potentials-based initial conditions $\mathbf{w_0}$ for the dynamic nodes and a set of time-constant boundary conditions $\mathbf{w_T}$ at the terminals
Decomposing matrix $\mathbf{A} = [\mathbf{I^{n_{p}\times n_{p}}}  \mathbf{A_{R}}$] and application in eq. \eqref{eq:KCLhere} leads to the following representation of the dynamic system
\begin{eqnarray}
\frac{d \mathbf{Z}}{dt} = -\mathbf{A_{R}} \mathbf{F_{R}} \label{eq:dynsys1} \\
\mathbf{F_{R}} =\Lambda(\mathbf{W_{R}}) \label{eq:actfunc1}\\
\mathbf{W_{R}} = \mathbf{A_{R}^{T}} \mathbf{w} \label{eq:potentials} \\
\mathbf{Z}   =  \mathbf{C}(\mathbf{w}) \label{eq:capacityeq}
\end{eqnarray}

Discretizing in time, using e.g. the explicit Euler method and combining with eq. \eqref{eq:dynsys1} - \eqref{eq:capacityeq} yields the following update rule 
\begin{equation}
\mathbf{w^{i+1,T}}=\mathbf{w^{i,T}}+\Delta t \mathbf{C}^{-1}(\Lambda(\mathbf{w^{i,T}}\mathbf{A_{R}})\mathbf{A_{R}^{T}})
\label{eq:euler}
\end{equation} where the right hand side $NN( \mathbf{w_{0},\mathbf{w_{T}}})$ is a neural network model as a function of the initial states $\mathbf{w_{0}}$ and terminal conditions $\mathbf{w_{T}}$ allowing to compute all states along the time trajectory of the dynamic process network. If we assume that the potentials at the process nodes and terminals are measurable and available from data over time, then the algorithm in Eq. \eqref{eq:euler} can be derived to calculate the trajectory of the process network through repeated forward passes through a neural network for each point in time $t$.

\begin{figure} [ht]
\centering
  \includegraphics*[width=\textwidth]{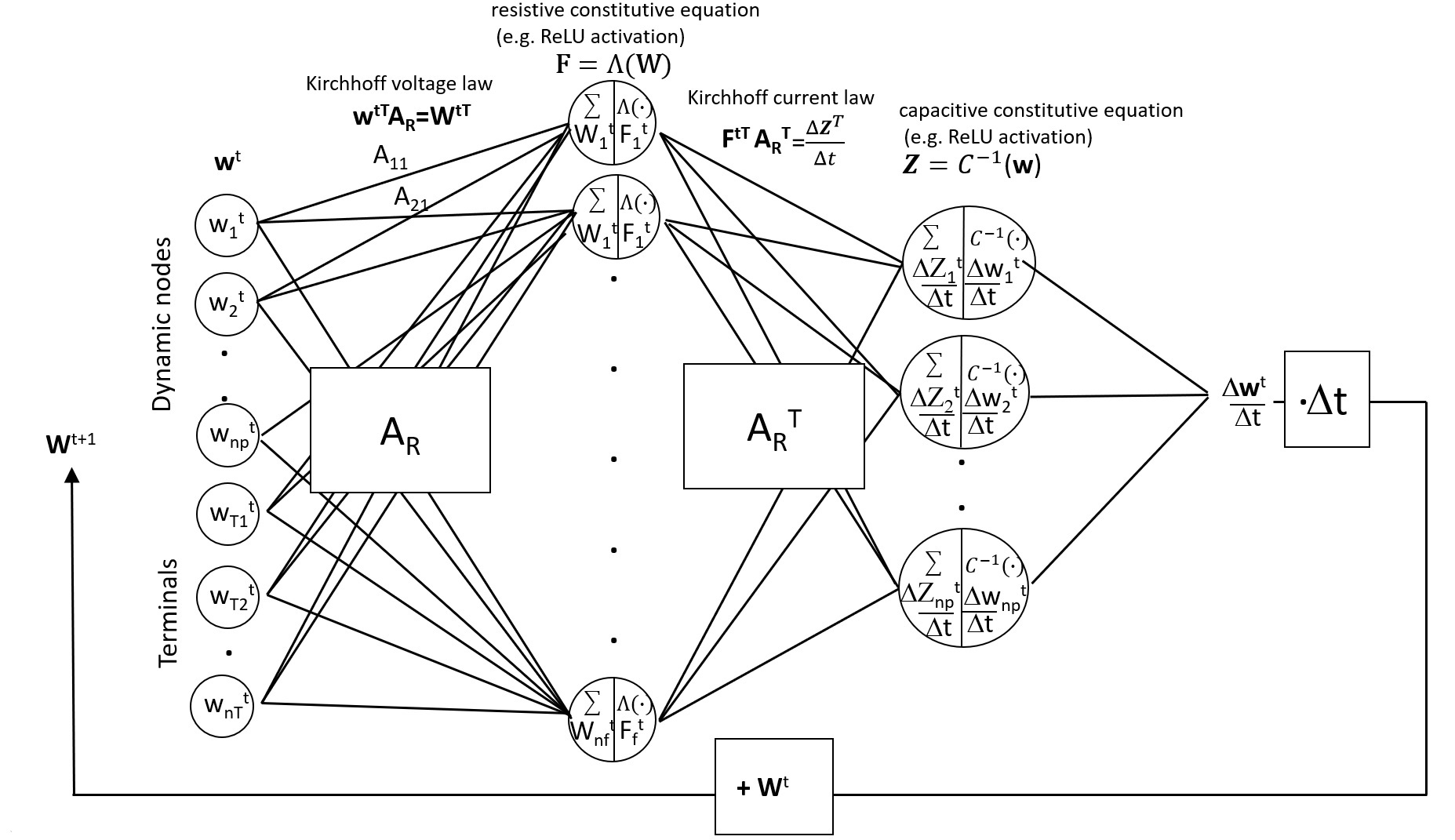}
  \caption{Neural network representation of the process network, with potentials $\mathbf{w}$ in the input layer, flows $\mathbf{F}$ at the hidden layer, and $\mathbf{\Delta w}$ at the output layer. The neural network represents the right hand side of a dynamic process network.}
\label{fig:NNpaper}
\end{figure}

In contrast to most ANN models, the resulting neural network is not fully connected but sparse and the connections between nodes represent the topology of the process network, i.e., edges have weights of zero where the process network has no physical connection between nodes. Further, the nodes of the neural network model and the functional values represent meaningful variables such as the potentials $\mathbf{w}$ in the input layer, the flows $\mathbf{F}$ at the hidden layer and the updates $\mathbf{\Delta w}$ at the output layer, see also Fig. \ref{fig:NNpaper}. Further, activation functions at the nodes are equivalent to constitutive equations. Depending on the expected relationship between e.g. flows and potentials for the resistive constitutive equation, they can be modeled with a linear activation function such as the identity $y=x$, a rectified linear model (ReLU), or even nonlinear functions such as $y=tanh(x)$. 

More complex non-linearities of the process network can be modeled through higher granularity for the different unit operations in the process system resulting into submodels of neural networks at the nodes connecting to the global network model rather than non-positive nonlinear functions (e.g. polynomials) as frequently used for process modeling.

\begin{figure} [ht]
\centering
  \includegraphics*[width=\textwidth]{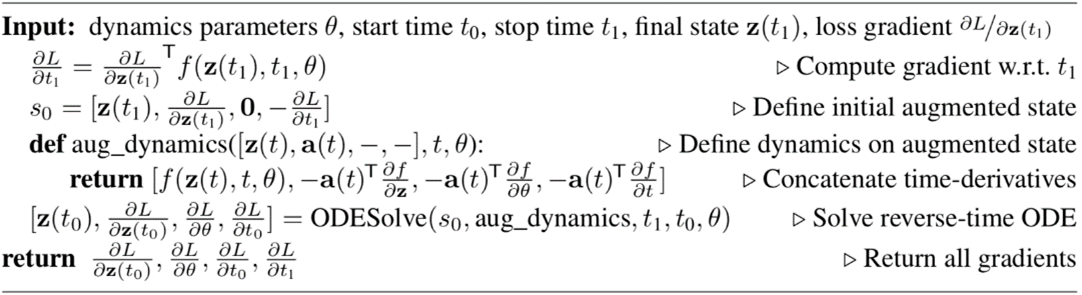}
  \caption{The algorithm describes backpropagation of gradients for the ODE initial value problem with subsequent observations. This algorithm lies in the heart of Neural ODEs from \cite{chenNODE}.}
\label{fig:alg_NODE}
\end{figure}

Until recently, time series neural network models of dynamic systems have frequently been trained using recurrent neural networks (RNN's) or Long Short Term Memory neural networks (LSTM) \cite{leeml}. Through the introduction of a new family of deep neural network models called neural ordinary differential equation systems (neural ODE's) by \cite{chenNODE}, deep neural networks in combination with classical ODE solvers can be used to train a neural network, see algorithm in Fig. \ref{fig:alg_NODE}. In their framework, instead of specifying a discrete sequence of hidden layers, the derivative is parameterized using a neural network with the help of adjoint differential equations.
The deep neural network models allow discretizing a dynamic system in time, in which each hidden layer of the neural network represents a particular time instance. As such, the approach has been used for learning differential equations from data. In the context of networked process systems whose dynamic behavior is typically modeled through classical differential equations and information for parameter estimation given through e.g. step testing or operational time-series data, this novel approach allows the process network formalism in this paper to be used as a thermodynamically consistent basis for applying ANN models in chemical engineering applications.

These models are especially powerful, when used in the context of time-series correlated data and allow learning differential equations from data. For process systems, dynamic models for control can be derived by applying a deep neural network if trained from e.g. operational step testing data.
Chen et al. \cite{chenNODE} show that for training of continuous-depth neural networks, reverse mode differentiation or backpropagation can best be carried out through an ODE solver method. The well-known adjoint method is suitable to calculate the gradients of a scalar-valued loss function L for neural network weight training when combined with a gradient decent method for back-propagation. From a process network perspective, learning the dynamic behavior of a set of time-series data is essentially equivalent to modeling the right-hand side of a differential equation system as in Eq. 1 through a neural network representation.

For a template of a neural network model with trainable weights, it would be ideal to have the process system’s topology already represented in the neural networks layers such that the (non-linear) relations between flows $\bf{F}$ and inventories $\bf{Z}$ and the potentials $\bf{w}$ are learned from data while inventory balances and their structure are pre-imposed. As shown in Fig. \ref{fig:NNpaper}, if potentials $\bf{w}$ of a network are measurable inputs, then continuity of the potentials allows relating them to differentials $\bf{W}$ and flow variables $\bf{F}$ through these relations (e.g. pressure differential drives convective flow). Further, flow variables $\bf{F}$ in the hidden layer relate to the inventory balance and allow calculation of the inventory differences $\mathbf{\Delta Z}$.
To represent the topology of a process network in a neural network model, some of the weights in the incident matrix A have to be set to zero resulting in a sparsely connected neural network where no physical connection is present. In the process of training the weights, these zero weights have to be pruned in training reducing the computational effort, preserving the process topology, and resulting in potentially explainable parameters for the remaining weights.

\section{An inventory system example}
\label{sec:pipeline}
\begin{figure} [ht]
\centering
  \includegraphics*[width=\textwidth]{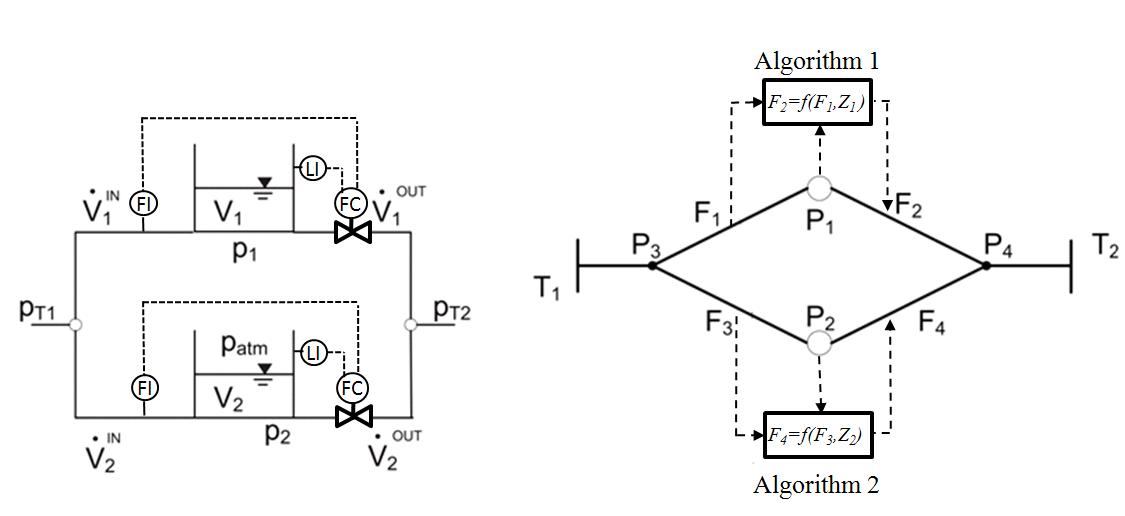}
  \caption{Graphical network representations: Problem specific representation on the left, a generalized representation on the right. The process control configuration with two control algorithms measuring the supply flows $F_{1}$ and $F_{3}$ and the inventories $Z_{1}$ and $Z_{2}$ at the nodes and controlling the demand flows $F_{2}$ and $F_{4}$.}
\label{fig:config}
\end{figure}

A small inventory control example shows how process networks optimize entropy dissipation and how control alters the objective function. The network consists of two connected pipelines where each pipeline flows through a cylindrical storage tank with volume $V_{j}$ open to the atmosphere, as shown in Fig. \ref{fig:config}. 

The pipeline flow is given as a lumped parameter representation introducing pressure potentials $w_{j}$ at the nodes. The relation between volumetric flow ${F_{i}}$ and pressure drop is considered a linear relationship for the resistive constitutive equations. The fluid volume $Z_{j}$ in the tank is connected to the level $h_{j}$ through $V_{j} = A_{j} h_{j}$ where $A_{j}$ is the cross-section of the tank resulting in a linear constitutive equation. We complete the model with the conservation laws for mass or, for constant density, the conservation of volume: \begin{eqnarray}
d{Z}_{1}/dt &=& F_{1}- F_{2} \label{eq:KCL1a} \\
d{Z}_{2}/dt &=& \dot{F}_{3}- F_{4} \label{eq:KCL2a} \\
{F}_{T1} &=& F_{1} + F_{3} \label{eq:KCL3a} \\
{F}_{T2} &=& F_{2} + F_{4} \label{eq:KCL4a}
 \end{eqnarray} Initial conditions for the tank volumes $Z_{0,i}$ have to be specified as well as boundary conditions at the terminals. The steady state of (\ref{eq:KCL1a}) - (\ref{eq:KCL4a}) can be found by integrating the differential equations.

The dynamic system given by the previous equations converges to the solution of the following optimization problem ($\frac{dZ_{1}}{dt}=\frac{dZ_{2}}{dt}=0$): 
\noindent
\begin{eqnarray}
\min_{\mathbf{w}} &  G^{*} = \sum_{i=1}^{4} \int_{0}^{W_{i}} F_{i} d W_{i} \label{eq:opt1}\\
s.t. & W_{1} = w_{T1} - w_{1} \\
		 & W_{2} = w_{1} - w_{T2} \\
		 & W_{3} = w_{T1} - w_{2} \\
		 & W_{4} = w_{2} - w_{T2} \\
     & F_{i} = K_{i} W_{i} \text{, } i=1..4 \\
     & Z_{i} = C_{i} w_{i} \text{, } i=1..2 \\
     & w_{T1} = \text{const, } w_{T2} = \text{const} \label{eq:optend}     
 \end{eqnarray}  Solving the optimization problem therefore corresponds to minimizing the power dissipated through viscous friction in the pipes subject to the conservation laws and boundary conditions.
 
The resulting potential function is quadratic
\begin{equation}
P(\mathbf{Z}) = \frac{1}{2} \mathbf{Z}^{T} \Phi_{Z} \mathbf{Z} - \mathbf{Z}^{T} \Phi_{Z,w_{T}} \mathbf{w_{T}} + \frac{1}{2} \mathbf{w_{T}}^{T} \Phi_{w_{T}} \mathbf{w_{T}} 
\end{equation} 
The equilibrium points follow from the first order conditions

\begin{displaymath}
Z_{1}^{*} = 1/C_{1} \frac{K_{1}w_{T1} + K_{3}w_{T2}}{K_{1} + K_{3}} \text{,  } Z_{2}^{*} = 1/C_{2} \frac{K_{2}w_{T1} + K_{4}w_{T2}}{K_{2} + K_{4}}
\end{displaymath}

To illustrate the effect of linear control, we consider inventory control on the flows $F_{2}$ and $F_{4}$. The control configuration and algorithms are displayed in Fig. \ref{fig:config}.

The optimization problem is as in equations \eqref{eq:opt1} - \eqref{eq:optend} including the two additional constraints adding the control laws \begin{eqnarray}
F_{2} = F_{1} - K_{C,1} (Z_{1} - Z_{1}^{c}) \\
F_{4} = F_{3} - K_{C,2} (Z_{2} - Z_{2}^{c})
\end{eqnarray} The potential function is derived as
\begin{eqnarray}
P = \int_{0}^{W_{1}} F_{1} d W_{1} + \int_{0}^{W_{2}} [F_{1} - K_{C,1} (Z_{1} - Z_{1}^{c})] d W_{2} \\
 + \int_{0}^{W_{3}} F_{3} d W_{3} + \int_{0}^{W_{4}} [F_{3} - K_{C,2} (Z_{2} - Z_{2}^{c})] d W_{4}
\end{eqnarray} Using $d W_{1} = - d W_{2} = - d w_{1}$ and $d W_{3} = - d W_{4} = - d w_{2}$ gives
\begin{equation}
P =  \int_{0}^{w_{1}} K_{C,1} (Z_{1} - Z_{1}^{c}) d w_{1} + \int_{0}^{w_{2}} [K_{C,2} (Z_{2} - Z_{2}^{c})] d w_{2}
\end{equation} Using the capacitive constitutive equations, the resulting potential function is quadratic
\begin{equation}
P(Z_{1},Z_{2}) = \frac{1}{2} \frac{K_{C,1}}{C_{1}} (Z_{1} - Z_{1}^{c})^{2} + \frac{1}{2} \frac{K_{C,2}}{C_{2}} (Z_{2} - Z_{2}^{c})^{2}
\end{equation} The equilibrium points trivially follow as $Z_{1,2} = Z_{1,2}^{c}$ from the first order conditions. Hence, convexity and a minimum follow from the second order conditions
\begin{eqnarray}
\frac{\partial^{2} P}{\partial Z_{1}^{2}} &=&  \frac{K_{C,1}}{C_{1}} > 0, \frac{\partial^{2} P}{\partial Z_{2}^{2}} =  \frac{K_{C,2}}{C_{2}} > 0 \\
\frac{\partial^{2} P}{\partial Z_{1}Z_{2}} &=&   0, \frac{\partial^{2} P}{\partial Z_{2}Z_{1}} =   0 
\end{eqnarray} for positive $\frac{K_{C,i}}{C_{i}}$. The resulting dynamic behavior for the process networks and control algorithms is given as the gradient system
\begin{eqnarray}
\frac{1}{C_{1}} \frac{d Z_{1}}{dt} = \frac{\partial P}{\partial Z_{1}}, \frac{1}{C_{2}} \frac{d Z_{2}}{dt} = \frac{\partial P}{\partial Z_{2}}
\end{eqnarray}

It is apparent that the value of the objective function as well as the flows of the dynamic simulation converge to the optimum determined through the optimization problem for arbitrary initial conditions. The constant inflow $\dot{V}_{T1}$ into the network divides itself into flows through the upper segments and lower segments choosing the path of least resistance.

We apply the formalism introduced for neural networks for the flow system and derive the weight matrices for the neural network from network theory. The Euler method based update law is given as

\begin{equation}
\mathbf{w^{i+1,T}}=\mathbf{w^{t,T}}+\Delta t \mathbf{ReLU}(\mathbf{ReLU}(\mathbf{w^{i,T}}\mathbf{A_{K}})\mathbf{A_{C}^{T}})
\label{eq:eulerpipes}
\end{equation} with $\mathbf{w^{t,T}} =[w_{1}^{t}, w_{2}^{t}, w_{T1}, w_{T2}]$ for time $t$ the resistive weights incident matrix $ \mathbf{A_{K}} = \mathbf{K A_{R}}$

$\begin{array}{ll}
	 \\
	\mathbf{A_{K}} = 
	
	\left[
	\begin{array}{cccc}
		 -K_{1} & 0 & K_{1} & 0 \\
		 K_{2}  & 0 & 0 & -K_{2} \\
		 0  & -K_{3} & K_{3} & 0 \\
		 0 & K_{4} & 0 &  -K_{4} \\
	\end{array}
	\right]
	\begin{array}{l} F_{1} \\ F_{2} \\ F_{3} \\ F_{4}  \end{array} &
\end{array}$ 

and the capacitive weights incident matrix $\mathbf{A_{C}} = \mathbf{C^{-1} A_{R}^{T}}$ 

$\begin{array}{ll}
	 \\
	\mathbf{A_{C}} = 
	
	\left[
	\begin{array}{cccc}
		 1/C_{1} & -1/C_{1} & 0 & 0 \\
		 0 	& 0 & 1/C_{2} & -1/C_{2} \\
		 0 & 0 & 0 & 0 \\
		 0 & 0 & 0 & 0 \\
	\end{array}
	\right]
	\begin{array}{l} P_{1} \\ P_{2} \\ T_{1} \\ T_{2}  \end{array} &
\end{array}$

where terminals $T_{1}$ and $T_{2}$ are not dynamic and hence not incident with the flows. 

\begin{figure} [ht]
\centering
  \includegraphics*[width=\textwidth]{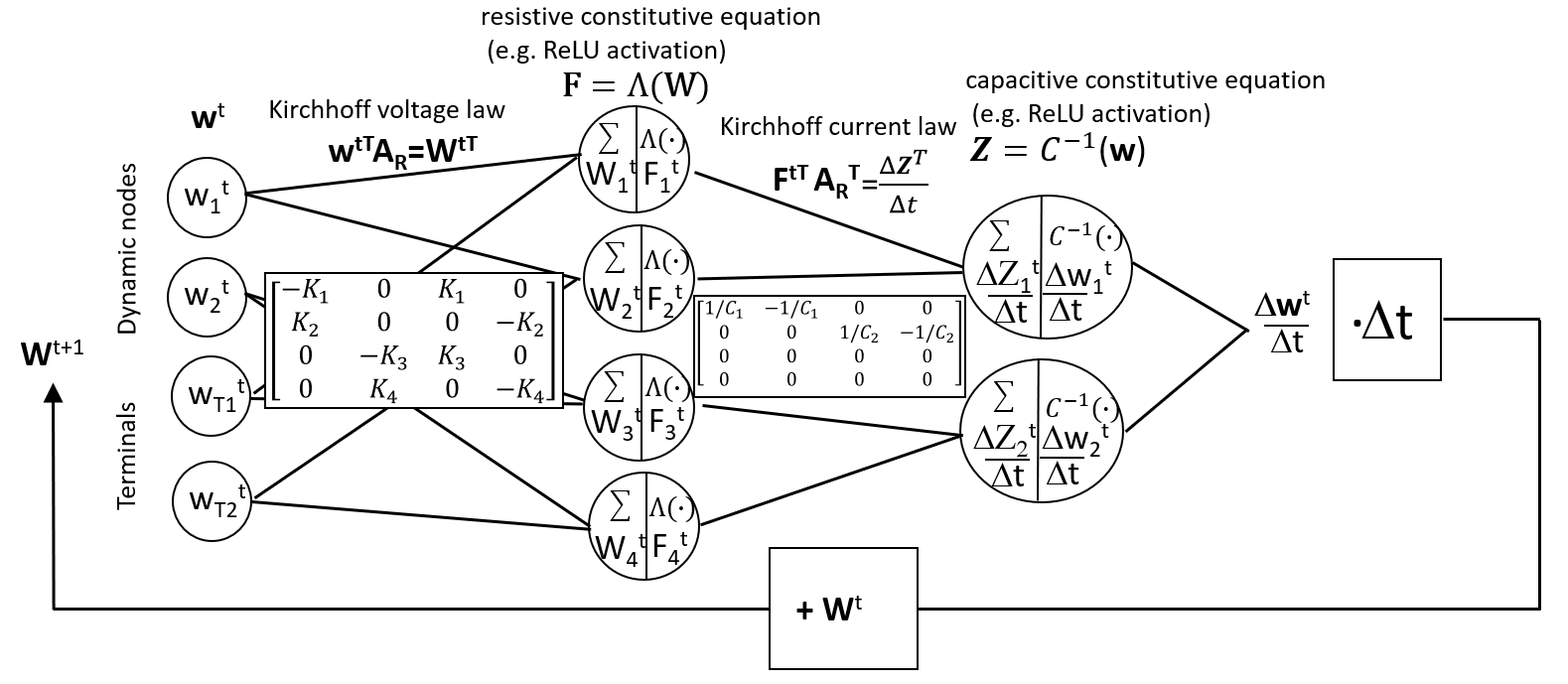}
  \caption{Neural network representation of the process network example, with potentials $\mathbf{w}$ in the input layer, flows $\mathbf{F}$ at the hidden layer, and potential differences $\mathbf{\Delta w}$ per time step at the output layer. Final weights after training as given in the connectivity matrices are equivalent to the differential equation system from which synthetic data was derived.}
\label{fig:exampleNN}
\end{figure}

The activation functions for the hidden nodes can be modeled as e.g. ReLUs $y(x)=ReLU(x)=max(0,x)$. The resulting neural network with 4 input nodes, 4 hidden nodes, and 2 output nodes is given in Fig. \ref{fig:exampleNN}.

This small dynamic flow and storage example shows how process networks can be modelled using neural ODE’s. 
The resulting linear classical ODE can be derived from the balances around the nodes $P_{1}$ and $P_{2}$ for $Z_{1}$ and $Z_{2}$, i.e., $\frac{dZ_{1,2}}{dt} = F_{1,3}-F_{2,4}$, constant terminal potentials $w_{T1}$ and $w_{T2}$, the flow constitutive equations $F_{i} = K_{i} W_{i} $ with i=1,..,4, the potential differences $W = w_{1} – w_{2}$ as in Eq. (4), and $w_i = C_i Z_i$

 \begin{eqnarray}
C_1^{-1}  \frac{dw_1}{dt} = -(K_1 - K_3 ) w_1 + K_1 w_T 1+ K_3 w_T2 \label{eq:diffw1} \\
C_2^{-1}  \frac{dw_2}{dt} = -(K_2 - K_4 ) w_2 + K_2 w_T1 + K_4 w_T2 \label{eq:diffw2} \\
 \end{eqnarray}

Discretization using the explicit Euler method from Eq. \ref{eq:eulerpipes} for demonstration purposes and using ReLU’s as activation functions leads the connectivity matrices AK and AC as given in Fig. \ref{fig:exampleNN}.

\begin{figure} [ht]
\centering
  \includegraphics*[width=\textwidth]{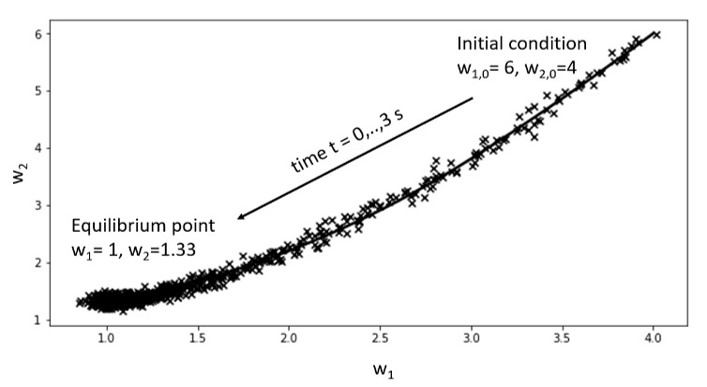}
  \caption{Time trajectory of dynamic potentials $w_1-w_2$ of inventory system after training the neural network. Synthetic time series data (marks) vs. neural network model (solid line).}
\label{fig:simtraj}
\end{figure}

The Neural ODE system was trained in Google Co-Lab using cloud GPU’s based on synthetic dynamic data with added Gaussian noise (5\%) from the corresponding ODE of Eqs. \ref{eq:diffw1} and \ref{eq:diffw2} with parameters $K_1=1$, $K_2=2$, $K_3=3$, $K_4=4$, $C_1=C_2=2$, and $w_{T1}=4$, $w_{T2}=0$. A Pytorch Python-based implementation was chosen with a time discretization using the explicit Euler method and time steps of $\Delta t = 0.02 $s. The process topology of the neural ODE was generated by pruning weights of non-incident edges, i.e., setting the corresponding weights to zero in every forward pass iteration during training. An adaptive deep learning stochastic gradient descent algorithm was applied for back-propagation to train the non-zero weights of the neural network. Training batches were selected through randomly choosing time periods within the synthetic data and 1000 training iterations carried out. The training run took T=17 sec and the neural network model matched the original data of the synthetic model where weights match with the original ODE (see Fig. \ref{fig:simtraj}) for the final $w_1-w_2$ time trajectory versus training data. 

The resulting neural network simulates the dynamic system even if initial conditions are chosen differently from the original training data, i.e., extrapolates versus the dynamic trajectory of the original data.
 
\section{Conclusions and Discussion}

We introduced a framework for analysis of optimality of networked process systems. We provide a systematic approach to derive optimality conditions for these systems. The objective function minimized by a process systems in its steady state and the influence of linear inventory control on the objective function is derived. This allows reshaping the objective function of a process network towards an engineering or economic objective. 
Using a graph theory based neural network representation for a process network allows using the topology of the system to be incorporated into the neural network through the incident matrices between edges and nodes. A sparsely connected neural network with meaningful topology allows preservation of e.g. inventory balances and a consistent potential field. Relationships between extensive quantities such as flows and inventories to intensive quantities, i.e., potentials are learned from data through training the remaining non-zero weights in the neural network through pruning. By applying classical ODE solvers and discretizing in time via, e.g., the explicit Euler method in combination with adjoint equations allows determining gradients through back-propagation with a deep learning stochastic gradient descent method. Neural ODE models for process systems networks have the potential to be used for model predictive control. Weight parameters become explainable and can be understood and updated when new data becomes available or the fundamental process structure is changed.


\begin{thebibliography}{18}
\expandafter\ifx\csname natexlab\endcsname\relax\def\natexlab#1{#1}\fi
\providecommand{\url}[1]{\texttt{#1}}
\providecommand{\href}[2]{#2}
\providecommand{\path}[1]{#1}
\providecommand{\DOIprefix}{doi:}
\providecommand{\ArXivprefix}{arXiv:}
\providecommand{\URLprefix}{URL: }
\providecommand{\Pubmedprefix}{pmid:}
\providecommand{\doi}[1]{\href{http://dx.doi.org/#1}{\path{#1}}}
\providecommand{\Pubmed}[1]{\href{pmid:#1}{\path{#1}}}
\providecommand{\bibinfo}[2]{#2}
\ifx\xfnm\relax \def\xfnm[#1]{\unskip,\space#1}\fi
\bibitem[Chen et~al.(2018)]{chenNODE}
\bibinfo{author}{Chen, R. T.~Q.}, \bibinfo{author}{Rubanova, Y.},
  \bibinfo{author}{Bettencourt, J.}, \bibinfo{author}{Duvenaud, D.},
  \bibinfo{year}{2018}.
\newblock \bibinfo{title}{Neural ordinary differential equations}.
\newblock \bibinfo{journal}{ArXiv 1806.07366}.

\bibitem[Doran et~al.(2017)]{XAI}
\bibinfo{author}{Doran, D.}, \bibinfo{author}{Schulz, S.},
  \bibinfo{author}{Besold, T.~R.}, \bibinfo{year}{2017}.
\newblock \bibinfo{title}{What does explainable ai really mean? a new
  conceptualization of perspectives}.
\newblock \bibinfo{journal}{ArXiv 1710.00794}.

\bibitem[Garcia and Morari(1982)]{garcia}
\bibinfo{author}{Garcia, C.~E.}, \bibinfo{author}{Morari, M.},
  \bibinfo{year}{1982}.
\newblock \bibinfo{title}{Internal model control. a unifying review and some new
  results}.
\newblock \bibinfo{journal}{Ind. Eng. Chem. Process Des.}
  \bibinfo{volume}{21}, \bibinfo{pages}{308--323}.

\bibitem[Hangos et~al.(1999)]{hangos}
\bibinfo{author}{Hangos, K.~M.}, \bibinfo{author}{Alonso, A.~A.},
  \bibinfo{author}{Perkins, J.~D.}, \bibinfo{author}{Ydstie, B.~E.},
  \bibinfo{year}{1999}.
\newblock \bibinfo{title}{Thermodynamic approach to the structural stability of
  process plants}.
\newblock \bibinfo{journal}{AICHE} \bibinfo{volume}{45}, \bibinfo{pages}{802--816}.

\bibitem[Jillson and Ydstie(2007)]{krjpaper}
\bibinfo{author}{Jillson, K.~R.}, \bibinfo{author}{Ydstie, B.~E.},
  \bibinfo{year}{2007}.
\newblock \bibinfo{title}{Process networks with decentralized inventory and
  flow control}.
\newblock \bibinfo{journal}{Journal of Process Control} \bibinfo{volume}{17},
  \bibinfo{pages}{399--413}.

\bibitem[Lee et~al.(2018)]{leeml}
\bibinfo{author}{Lee, J.~H.}, \bibinfo{author}{Shina, J.},
  \bibinfo{author}{Realff, M.~J.}, \bibinfo{year}{2018}.
\newblock \bibinfo{title}{Machine learning: Overview of the recent progresses
  and implications for the process systems engineering field}.
\newblock \bibinfo{journal}{Computers and Chemical Engineering}
  \bibinfo{volume}{114}, \bibinfo{pages}{111--121}.

\bibitem[Luenberger(1979)]{luenberger79}
\bibinfo{author}{Luenberger, D.~G.}, \bibinfo{year}{1979}.
\newblock \bibinfo{title}{Introduction to Dynamic Systems: Theory, Models, and
  Applications}.
\newblock \bibinfo{publisher}{Wiley, New York}.

\bibitem[Marquardt et~al.(2010)]{ontocape}
\bibinfo{author}{Marquardt, W.}, \bibinfo{author}{Morbach, J.},
  \bibinfo{author}{Wiesner, A.}, \bibinfo{author}{Yang, A.},
  \bibinfo{year}{2010}.
\newblock \bibinfo{title}{OntoCAPE - A Re-Usable Ontology for Chemical Process
  Engineering}.
\newblock \bibinfo{publisher}{Springer, New York}.

\bibitem[Maxwell(1892)]{maxwell}
\bibinfo{author}{Maxwell, J.~C.}, \bibinfo{year}{1892}.
\newblock \bibinfo{title}{A Treatise on Electricity and Magnetism}.
\newblock \bibinfo{publisher}{Oxford University Press}.

\bibitem[Ning and You(2019)]{fengqiML}
\bibinfo{author}{Ning, C.}, \bibinfo{author}{You, F.}, \bibinfo{year}{2019}.
\newblock \bibinfo{title}{Optimization under uncertainty in the era of big data
  and deep learning: When machine learning meets mathematical programming}.
\newblock \bibinfo{journal}{Computers and Chemical Engineering}
  \bibinfo{volume}{125}, \bibinfo{pages}{434--448}.

\bibitem[Oster et~al.(1971)]{oster2}
\bibinfo{author}{Oster, G.~F.}, \bibinfo{author}{Perelson, A.~L.},
  \bibinfo{author}{Katchalsky, A.~Q.}, \bibinfo{year}{1971}.
\newblock \bibinfo{title}{Network thermodynamics}.
\newblock \bibinfo{journal}{Nature} \bibinfo{volume}{234},
  \bibinfo{pages}{393--399}.

\bibitem[Peusner(1986)]{peusner}
\bibinfo{author}{Peusner, L.}, \bibinfo{year}{1986}.
\newblock \bibinfo{title}{Studies in Network Thermodynamics}.
\newblock \bibinfo{publisher}{Elsevier, Amsterdam}.

\bibitem[Prigogine(1947)]{prigogine1947}
\bibinfo{author}{Prigogine, I.}, \bibinfo{year}{1947}.
\newblock \bibinfo{title}{Etude thermodynamique des phenomenes irreversibles}.
\newblock Ph.D. thesis, \bibinfo{school}{Liege}.

\bibitem[Sch{\"a}fer et~al.(2020)]{schaefer}
\bibinfo{author}{Sch{\"a}fer, P.}, \bibinfo{author}{Caspari, A.},
  \bibinfo{author}{Schweidtmann, A.~M.}, \bibinfo{author}{Vaupel, Y.},
  \bibinfo{author}{Mhamdi, A.}, \bibinfo{author}{Mitsos, A.},
  \bibinfo{year}{2020}.
\newblock \bibinfo{title}{The potential of hybrid mechanistic/data-driven
  approaches for reduced dynamic modeling: Application to distillation
  columns}.
\newblock \bibinfo{journal}{Chemie Ingenieur Technik} \bibinfo{volume}{92},
  \bibinfo{pages}{1910--1920}.

\bibitem[Skogestad(2004)]{skogestad}
\bibinfo{author}{Skogestad, S.}, \bibinfo{year}{2004}.
\newblock \bibinfo{title}{Near-optimal operation by self-optimizing control:
  from process control to marathon running and business systems}.
\newblock \bibinfo{journal}{Computers and Chemical Engineering}
  \bibinfo{volume}{29}, \bibinfo{pages}{127--137}.

\bibitem[Venkatasubramanian(2019)]{venkat}
\bibinfo{author}{Venkatasubramanian, V.}, \bibinfo{year}{2019}.
\newblock \bibinfo{title}{The promise of artificial intelligence in chemical
  engineering: Is it here, finally?}
\newblock \bibinfo{journal}{AICHE} \bibinfo{volume}{65},
  \bibinfo{pages}{466--478}.

\bibitem[Wartmann and Ydstie(2009)]{ADCHEM2009}
\bibinfo{author}{Wartmann, M.~R.}, \bibinfo{author}{Ydstie, B.~E.},
  \bibinfo{year}{2009}.
\newblock \bibinfo{title}{Network-based analysis of stability,optimality of
  process networks}.
\newblock In: \bibinfo{booktitle}{Proceedings of International Symposium on
  Advanced Control of Chemical Processes}. pp. \bibinfo{pages}{197--202}.

\bibitem[Ydstie and Alonso(1997)]{ya97}
\bibinfo{author}{Ydstie, B.~E.}, \bibinfo{author}{Alonso, A.~A.},
  \bibinfo{year}{1997}.
\newblock \bibinfo{title}{Process systems and passivity via the
  clausius-planck inequality}.
\newblock \bibinfo{journal}{Systems \& Control Letters} \bibinfo{volume}{30},
  \bibinfo{pages}{253--264}.

\end{thebibliography}
\end{document}